%% file: root.tex
\documentclass[letterpaper, 10 pt, conference]{ieeeconf}  

\IEEEoverridecommandlockouts                              

\usepackage[subtle]{savetrees}

\usepackage{cite}
\usepackage{balance}
\let\labelindent\relax
\usepackage{enumitem}

\usepackage{multicol}
\usepackage[bookmarks=true]{hyperref}
\usepackage[dvipsnames]{xcolor} 
\usepackage{booktabs} 
\usepackage{lipsum}
\usepackage{amsfonts}
\usepackage{amsmath}
\usepackage{xspace}
\usepackage[font=small,labelfont=bf]{caption} 

\newcommand{\algname}{Interactive Language\xspace}
\newcommand{\envname}{Language-Table\xspace}

\newcommand{\DCOLLECT}{\mathcal{D}_{\operatorname{collect}}}\xspace
\newcommand{\DTRAINING}{\mathcal{D}_{\operatorname{training}}}\xspace

\definecolor{SimColor}{rgb}{1.0,0.2,0.0}
\definecolor{RealColor}{rgb}{0.0,0.2,1.0}
\definecolor{LinkBlue}{rgb}{0.3,0.5,1.0}
\usepackage{pifont}
\newcommand{\cmark}{\textcolor[HTML]{59a14f}{\ding{51}}}
\newcommand{\xmark}{\textcolor[HTML]{e15759}{\ding{55}}}
\newcommand{\link}[1]{\color{LinkBlue}{\href{#1}{#1}}}

\usepackage{lipsum}
\usepackage{graphicx}
\usepackage{textcomp}
\usepackage{amsmath}

\title{\LARGE \bf
Interactive Language: Talking to Robots in Real Time
}

\makeatletter
\newcommand{\linebreakand}{%
  \end{@IEEEauthorhalign}
  \hfill\mbox{}\par
  \mbox{}\hfill\begin{@IEEEauthorhalign}
}
\makeatother

\author{
\vspace{-0.7cm}
\authorblockN{Corey Lynch, Ayzaan Wahid, Jonathan Tompson}
\linebreakand
\vspace{0.1cm}
\authorblockN{Tianli Ding, James Betker, Robert Baruch, Travis Armstrong, Pete Florence}
\linebreakand
\authorblockN{Robotics at Google}
\\[-5.0ex]
}

\begin{document}

\maketitle
\thispagestyle{empty}
\pagestyle{empty}

\vspace*{-6mm}
\begin{abstract}
We present a framework for building interactive, real-time, natural language-instructable robots in the real world, and we open source related assets (dataset, environment,  benchmark, and policies).
Trained with behavioral cloning on a dataset of hundreds of thousands of  language-annotated trajectories, a produced policy can proficiently execute an order of magnitude more commands than previous works: specifically we estimate a 93.5\% success rate on a set of 87,000 unique natural language strings specifying raw end-to-end visuo-linguo-motor skills in the real world.
We find that the same policy is capable of being guided by a human via real-time language to address a wide range of precise long-horizon rearrangement goals, e.g. ``\textit{make a smiley face out of blocks}".
The dataset we release comprises nearly 600,000 language-labeled trajectories, an order of magnitude larger than prior available datasets.
We hope the demonstrated results and associated assets enable further advancement of helpful, capable, natural-language-interactable robots. See videos at \link{https://interactive-language.github.io}.

\end{abstract}


\section{Introduction}

\input{v1_intro}

\begin{figure*}[h!]
\centering
\includegraphics[width=\textwidth]{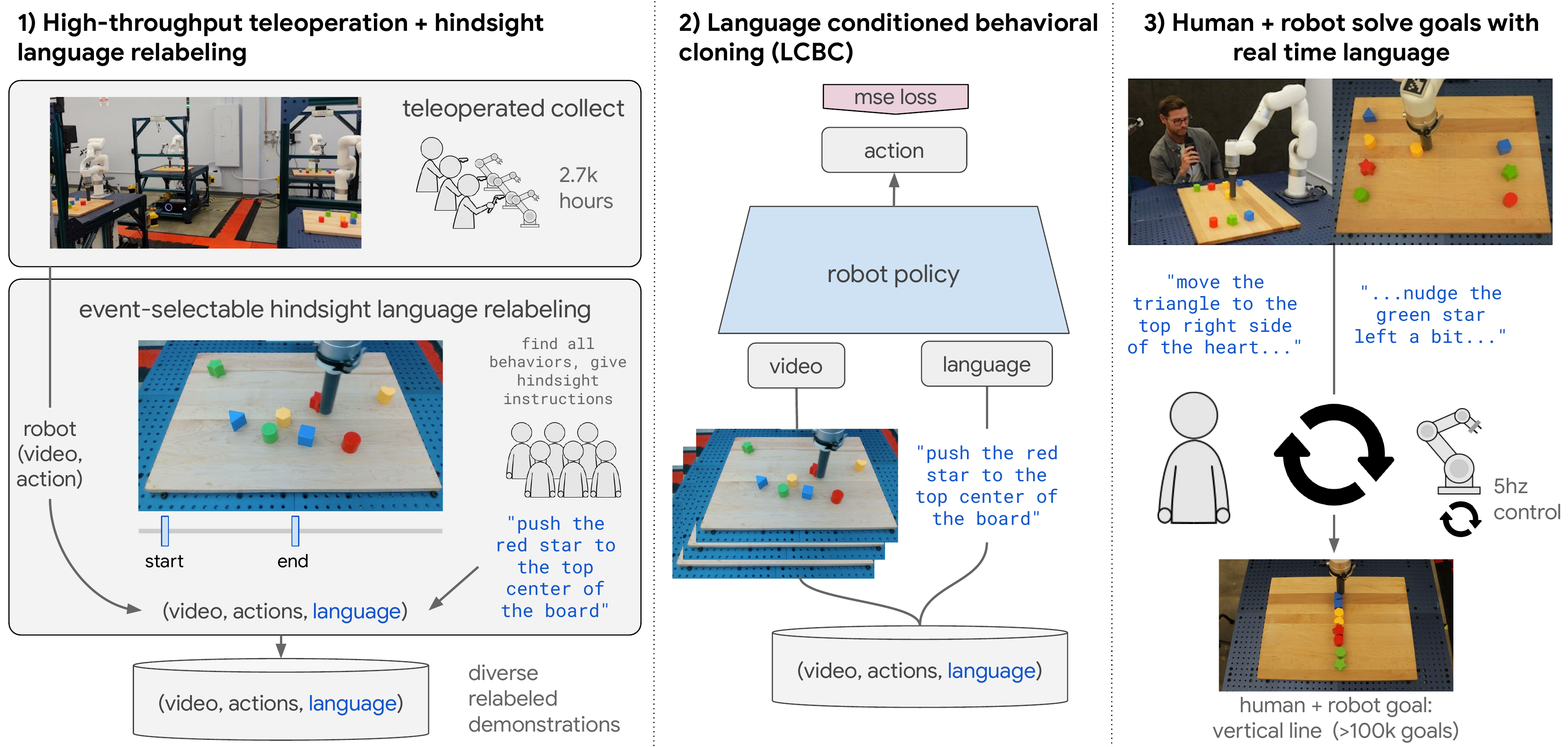}
\caption{\textbf{Interactive Language: a large scale robot imitation learning framework for real-time language.}  Stage 1: First, high throughput robot data collection with multiple operators. Post-collection, relabel robot video and actions into language conditioned demonstrations using event-selectable hindsight relabeling. Stage 2: do simple language conditioned behavioral cloning. Stage 3: Human guides a single learned policy in real-time using natural language to accomplish hundreds of thousands of goals.}
\label{fig:system}
\vspace{-5mm}
\end{figure*}

\textbf{Contributions}. Our primary contributions include (i) \algname, a framework for producing real world 
robots that can capably receive interactive open vocabulary language conditioning in real-time\footnote{For the scope of this paper, by real-time we mean new language conditioning can occur in the ``blink of an eye'', i.e. approximately 3 Hz \cite{kwon2013high} or greater.} while performing continuous-control visuomotor manipulation.
\algname combines existing techniques, together with novel components like event-selectable hindsight relabeling, to define a simple and scalable recipe for learning large repertoires of natural-language-conditionable skills.
(ii) We use this system to present and study the setting of \textit{interactive language guidance}, 
showing that the combination of real-time language feedback and a low-level language-conditionable policy can address 
long-horizon manipulation goal states in a tabletop rearrangement setting. 
(iii)  To facilitate future research in this domain, we release \textit{\envname}, a dataset and simulated multitask imitation learning benchmark. 
With nearly 600,000 diverse demonstrations across simulation and the real world, \envname is, to our knowledge, the largest natural language conditioned imitation learning dataset of its kind by an order of magnitude (Table~\ref{tab:data-comparison}).

\section{Related work}

\textbf{From single-task imitation to multi-task and language conditioning}.
Imitation learning (see review \cite{osa2018algorithmic}), the perspective we adopt in this work, provides a simple and stable way for robots to acquire behaviors from human expert demonstrations.
While historically imitation learning has been applied to individual tasks from instrumented state \cite{pomerleau1989alvinn, schaal2003computational, khansari2011learning, schaal2005learning},
the desire for more general purpose robots has motivated study into policies capable of learning multiple skills at once from more generic on-board sensory observations like RGB pixels \cite{caruana1997multitask, levine2016end,zhang2018deep}.
To condition multiple learned behaviors, prior setups have relied on discrete one-hot task identifiers \cite{rahmatizadeh2018vision}, which can be difficult to scale to many tasks, or goal images \cite{ding2019goal,lynch2020learning,chebotar2021actionable}, which can be impractical to provide in real world scenarios. Alternatively, a long history of prior work in broader AI research \cite{winograd1972understanding,branavan2009reinforcement, tellex2011understanding, chen2011learning, matuszek2013learning,hill2019emergent,lynch2020language} has sought a more convenient form of specification in the form of natural language conditioning (survey \cite{luketina2019survey}), with some results on physical robots \cite{stepputtis2020language,jang2022bc,nair2022learning,shridhar2022cliport}. This focus has yielded many varied and impressive approaches to tackling the grounding problem\cite{winograd1972understanding, hermann2017grounded}---learning to relate language to one's embodied observations and actions.
However, in both simulation and the real world, instruction-following robots rarely leverage the full capabilities of continuous control, instead employing simplified, parameterized action spaces \cite{wang2019reinforced, shridhar2022cliport, team2021creating, hill2019emergent}. Furthermore, once provided, language conditioning is typically presumed fixed over robot execution \cite{stepputtis2020language, nair2022learning, jang2022bc, ahn2022can}, with little opportunity for subsequent interaction by the instructor.
Our work, in contrast, studies the first combination, to our knowledge, of real-time natural language guidance of a physical robot engaged in continuous visuomotor manipulation.

\textbf{Interactively guiding robot behavior with language}. Our work exists in a larger setting of humans modifying or correcting the behavior of autonomous agents \cite{tellex2014asking}, historically addressed in forms like teleoperation \cite{rakita2018autonomous,kelly2019hg,spencer2020learning}, kinesthetic teaching \cite{kormushev2011imitation}, or sparse human preference feedback \cite{christiano2017deep}. Certain works have studied language as a means of correction, but typically do so under simplifying assumptions that we relax in the current work. 
For example, \cite{broad2016towards}, \cite{karamcheti2022lila}, \cite{co2018guiding}, and \cite{sharma2022correcting} study language corrections, but under the respective simplifying assumptions of hand-defined optimization for grounding, undivided operator attention, paired iterative corrections at training time, and presumed access to motion planners and task cost functions.
Additionally, to the best of our knowledge, none of these works support multiple-Hz iterative specification over the course of execution.
Closest to our approach is \cite{lynch2020language} and \cite{team2021creating}, which study language-interactive agents learned via imitation, but entirely in simulation and under varying degrees of actuation realism.
In contrast to these prior studies, our work learns real-time natural language policies end-to-end from RGB pixels to continuous control outputs with a simple behavioral cloning objective \cite{Pomerleau_behavior_cloning}, and applies them to contact-rich real-world manipulation tasks. 


\textbf{Scaling real world imitation learning.} 
One of the largest bottlenecks in robot imitation is often simply the amount of diverse robot data made available to learning \cite{rahmatizadeh2018vision,zhang2018deep,jang2022bc}. 
Many multi-task imitation learning frameworks determine the set of tasks to be learned upfront \cite{osa2018algorithmic,stepputtis2020language,jang2022bc,shridhar2022cliport,ahn2022can}. While this may simplify collection conceptually, it also often requires that reset protocols and success criteria be designed manually for each behavior. 
Another challenge particular to large scale multi-operator collections is that typically not all data can be considered optimal \cite{mandlekar2021matters, yang2021trail}, often requiring manual post-hoc success filtering \cite{jang2022bc, ahn2022can}. These per-task manual efforts have historically been difficult to scale to a large and diverse task setting, like the one studied in this work.
We sidestep both these scaling concerns by instead having operators continuously teleoperate long-horizon behaviors, with no requirements on low level task segmentation or resets \cite{lynch2020learning,gupta2019relay,lynch2020language} and then leverage after-the-fact crowdsourced language annotation \cite{lynch2020language, nair2022learning}. In contrast to the ``random window" relabeling explored in \cite{lynch2020language}, we give annotators precise control over the start and end of behaviors they are annotating, which we find in practice better aligns relabeled training data to the actual commands given at test time.


\section{Problem Setup}
Our goal is to train a conditional policy, $\pi_{\theta}(a | s, l)$, parameterized by $\theta$, which maps from observations $s \in \mathcal{S}$ and human-provided language $l \in \mathcal{L}$ to actions $a \in \mathcal{A}$ on a physical robot. 
In particular we are interested in \textit{open-vocabulary language-conditioned visuomotor policies}, in which the observation space contains high-dimensional RGB images, e.g. $\mathcal{S} = \mathbb{R}^{H \times W \times C}$, and where language conditioning $\mathcal{L}$ has no predefined template, grammar, or vocabulary. We are also particularly interested in allowing humans to interject new language $\mathcal{L}$ at any time, at the natural rate of the visuo-linguo-motor policy. Each commanded $l$ encodes a distribution of achievable goals $g^{short} \in \mathcal{G}^{short}$ in the environment.
Note that humans may generate a new language instruction $l$ based on their own perception of the environment, $s^H \in \mathcal{S}^H$, which may differ substantially from the robot's $s \in \mathcal{S}$ (e.g. due to viewpoint, self-occlusion, limited observational memory, etc.).
As in prior works \cite{lynch2020language}, we treat natural-language-conditioned visuomotor skill learning as a contextual imitation learning problem \cite{osa2018algorithmic}. As such, we acquire an offline dataset $\mathcal{D}$ containing pairs of valid demonstrations and the conditions they resolve $\{(\tau, l)_i\}^{\mathcal{D}}_{i=0}$. Each $\tau_i$ is a variable-length trajectory of robot observations and actions $\tau_i = [(s_0, a_0),(s_1, a_1), ..., (s_T)]$, and each $l_i$ describes the full trajectory as a second-person command.



\section{Interactive Language: Methods and Analysis}

First we introduce \textit{\algname}, summarized in Figure~\ref{fig:system}, a simple and generically applicable imitation learning framework for training real-time natural-language-interactable robots. \algname combines a scalable method for collecting varied, real world language-conditioned demonstration datasets, with straightforward language conditioned behavioral cloning (LCBC).

\subsection{Data Collection}

\begin{table}[]
\setlength\tabcolsep{4pt}
\begin{tabular}{|l|l|l|l|}
\hline
                                                                  & \textbf{Has contact} & \textbf{\begin{tabular}[c]{@{}l@{}}Object/location\\-directed\\ instructions\end{tabular}} & \textbf{\begin{tabular}[c]{@{}l@{}}Compound\\ instructions\end{tabular}} \\ \hline
Random window \cite{lynch2020language, nair2022learning}                                                     & 86\%    & 47\%                                                                        & 16\%                                                          \\ \hline
\begin{tabular}[c]{@{}l@{}}Event-selectable (ours)\end{tabular} & 91\%    & 83\%                                                                        & $<$ 1\%                                                         \\ \hline
Real test instructions                                            & 89\%     & 84\%                                                                       & $<$ 1\%                                                          \\ \hline
\end{tabular}
\caption{\textbf{Which relabeling strategy aligns best with test-time language?}}
\label{tab:hindsight-strategies}
\end{table}

\begin{table}[]
\begin{tabular}{|ll|}
\hline
\multicolumn{2}{|l|}{\textbf{Real-World Data Collection}}                                                                                    \\ \hline
\multicolumn{1}{|l|}{Total robots}                                                                                & 4     \\ \hline
\multicolumn{1}{|l|}{Total teleoperators}                                                                         & 10    \\ \hline
\multicolumn{1}{|l|}{Total episodes}                                                                              & 16.4k \\ \hline
\multicolumn{1}{|l|}{\begin{tabular}[c]{@{}l@{}}Average episode length  (minutes)\end{tabular}}                  & 9.9   \\ \hline
\multicolumn{1}{|l|}{\begin{tabular}[c]{@{}l@{}}Total hours of collect time\end{tabular}}                      & 2.7k  \\ \hline
\multicolumn{2}{|l|}{\textbf{Hindsight Relabeling}}                                                                       \\ \hline
\multicolumn{1}{|l|}{\begin{tabular}[c]{@{}l@{}}Total crowdsourced annotators\end{tabular}}                    & 64    \\ \hline
\multicolumn{1}{|l|}{\begin{tabular}[c]{@{}l@{}}Total relabeled demonstrations obtained\end{tabular}}           & 299k  \\ \hline
\multicolumn{1}{|l|}{\begin{tabular}[c]{@{}l@{}}Total unique relabeled  instructions\end{tabular}}               & 87k   \\ \hline
\multicolumn{1}{|l|}{\begin{tabular}[c]{@{}l@{}}Average relabeled  demonstration length (seconds)\end{tabular}} & 5.8   \\ \hline
\multicolumn{1}{|l|}{\begin{tabular}[c]{@{}l@{}}Total number of hours of relabeled demonstrations obtained\end{tabular}} & 488   \\ \hline
\multicolumn{1}{|l|}{\begin{tabular}[c]{@{}l@{}}Total instruction hours / Collect hours\end{tabular}} & 18.06\%   \\ \hline
\end{tabular}
\caption{\textbf{Statistics: real-world collection and relabeling}. This data snapshot went into training and is a subset of the full \envname data.}
\label{tab:data-stats}
\vspace{-6mm}
\end{table}

\textbf{High throughput raw data collection}. 
\algname adopts purposefully minimal collection assumptions to maximize the flow of human demonstrated behavior to learning.
Operators teleoperate a variety of long-horizon behaviors constantly, without low-level task definition, segmentation, or episodic resets.
This strategy shares assumptions with ``play" collection \cite{lynch2020learning}, but
additionally guides collect towards temporally extended low-entropy states like lines, shapes, and complex arrangements.
Each collect episode lasts $\sim$10 minutes before a break, and is guided by multiple randomly chosen long-horizon prompts $p \in \mathcal{P}$ (e.g. ``\textit{make a square shape out of the blocks}"), drawn from the set of target long-horizon goals, which teleoperators are free to follow or ignore. We do not assume all of the data collected for each prompt $p$ is 
optimal (each $p$ is discarded after collecting).
In practice, our collection includes many inevitable edge cases that might otherwise require data cleaning, e.g. solving for the wrong $p$ or knocking blocks off table. We log all of these cases and incorporate them later on as training data.
Concretely, this collect procedure yields a \textit{semi-structured, optimality-agnostic} collection $\DCOLLECT = \{\tau_i\}^{\DCOLLECT}_{i=0}$. 
The purpose of $\DCOLLECT$ is to provide a sufficiently diverse basis for crowdsourced hindsight language relabeling \cite{lynch2020language,nair2022learning}, described next. 

\textbf{Event-selectable hindsight relabeling}. We convert $\DCOLLECT$ into 
natural language conditioned demonstrations $\DTRAINING = \{(\tau, l)_i\}^{\DTRAINING}_{i=0}$, using a new variant of hindsight language relabeling \cite{lynch2020language} we call ``Event-Selectable Hindsight Relabeling" (Fig.\ref{fig:system}, left).
Previous ``random window" relabeling systems \cite{lynch2020language, nair2022learning} have at least two drawbacks: each random window is not guaranteed to contain ``usefully describable" actions, and random window lengths must be determined upfront as a sensitive hyperparameter.
We instead ask annotators to watch the full collect video, then find $K$ coherent behaviors ($K=24$ in our case).
Annotators have the ability to mark the start and end frame of each behavior, and are asked to phrase their text descriptions as natural language commands. 
In Table~\ref{tab:hindsight-strategies}, we compare event-selectable relabeling to prior ``random window" relabeling on a subset of our training data. 
We find that while both strategies tend to describe contact-rich behaviors, our analysis suggests event-selectable relabeling yields more well-matched data: fewer complex compound instructions, and more compositionally directed instructions.


\textbf{Throughput and bottleneck analysis}.
Here, we share some insights gained from scaling our robot collect and hindsight relabeling operation.
See statistics on our collected data in Table~\ref{tab:data-stats}. 
We find, perhaps surprisingly, that the main bottleneck in our data operation is \textit{not} robot teleoperation but rather the crowdsourced language annotation that follows, with 18.06\% of the raw data having undergone annotation prior to model training (5.5x as much unlabeled collected data as annotated data). This is true even though there are 16x as many hindsight annotators as robots. Bottlenecks like this may be addressed by exploiting language-free co-training \cite{lynch2020language}, or by simply continuing to horizontally scale crowdsourced annotators.

\begin{figure}[h!]
    \vspace{-0.5cm}
    \noindent
    \centering
    \includegraphics[width=\linewidth]{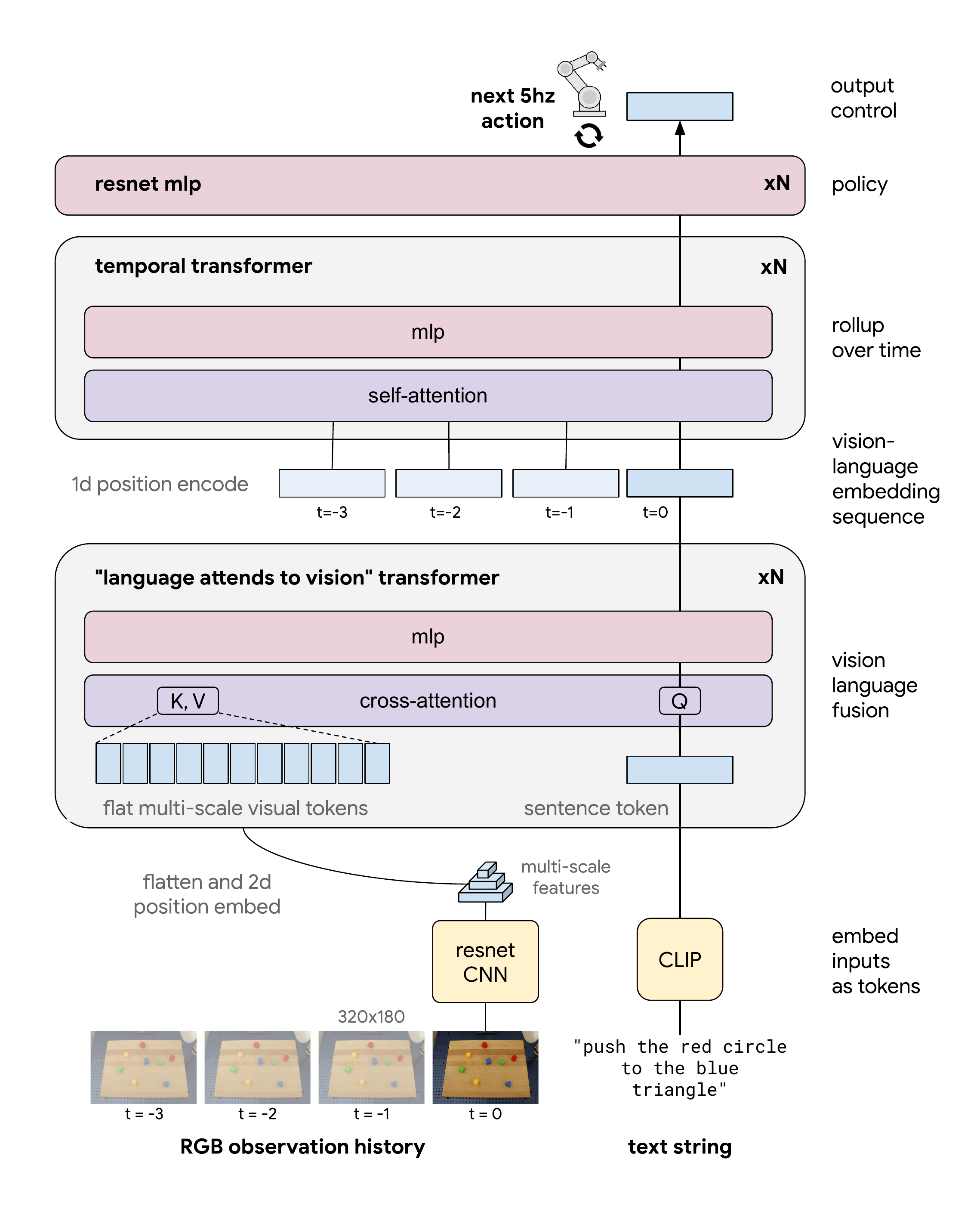}
    \vspace{-0.9cm}
    \caption{\textbf{LAVA: our transformer-based architecture for language conditioned visuomotor control.}}
    \label{fig:arch}
\vspace{-3mm}
\end{figure}

\subsection{Policy Learning}

\textbf{Transformer-based agent architecture}.
In Figure~\ref{fig:arch}, we describe our transformer-based \cite{vaswani2017attention} neural network policy architecture, mapping from video and text to continuous actions, which we refer to as LAVA (``Language Attends to Vision to Act").
Each training example consists of $(s, a, l)_i \sim \DTRAINING$, where
$s \in \mathbb{R}^{\operatorname{seqlen}\times640\times320\times3}$ is RGB observation history.
We ConvNet-process each frame in the video $s$ to obtain multi-scale visual features (features at multiple resolutions). The first two layers are Imagenet-pretrained ResNet \cite{deng2009imagenet, he2016deep}. $l$ is embedded using a pretrained CLIP text encoder \cite{radford2021learning}, which is finetuned on our in-domain data, but remains fixed during policy training. 
We fuse visual and lingual information using a ``Language-Attends-to-Vision" transformer block, which performs 
cross-attention with language acting as query, and flattened multi-scale visual tokens acting as keys and values.
This operation is applied to each image, and the sequence output is fed to a temporal prenorm \cite{xiong2020layer} transformer, which is average pooled and fed to a deep residual multi-layer perceptron (MLP), 
outputting the predicted next action $a$. 

\textbf{Training}.
We train our policies with a standard supervised language conditioned behavioral cloning (LCBC) objective.
While we expect that more complex loss functions or policy classes may acquire even better results, all the policies we present were trained as deterministic policies with a simple mean squared error loss: $\underset{\theta}{\min} \sum_{(s,a,l) \sim \DTRAINING}  ||a - \pi_{\theta}(s,\;\; l)||_2^2$, e.g. as in \cite{deepIL_2018,jang2022bc}.

\section{Language-Table: Datasets and Environment}
\label{sec:lang-table}
To facilitate further research in language-conditioned visuomotor learning, we release \textit{Language-Table}, 
which consists of (i) a suite of datasets and (ii) a simulated multi-task language conditioned control environment and benchmark.

\textbf{Dataset}. \envname provides our human-relabeled $\DTRAINING$ and the underlying human-teleoperated $\DCOLLECT$, both in simulation and the real world. 
The $\DTRAINING$ real and sim datasets are highlighted in Table \ref{tab:data-comparison} -- an order of magnitude larger than comparable, previously-available datasets.

\begin{table}[]
  \setlength\tabcolsep{5pt}
  \centering
  \begin{tabular}[b]{@{}lccccc@{}}
  \toprule  
          & \multicolumn{1}{c}{\#}      & \multicolumn{1}{c}{\#}  &  Physical     &       &           \\    
  Dataset & Traj. (k)      & Unique (k) &  Actions     & Real  & Available \\
  \midrule
  \multicolumn{3}{@{}l}{\textit{Episodic Demonstrations}} \\
  BC-Z \cite{jang2022bc} &   {\color{RealColor}25} & {\color{RealColor}0.1} & \cmark & \cmark & \cmark \\
  SayCan \cite{ahn2022can} & {\color{RealColor}68} & {\color{RealColor}0.5} & \cmark & \cmark & \xmark \\
  Playhouse \cite{team2021creating} & {\color{SimColor}1,097} & {\color{SimColor}779} & \xmark & \xmark & \xmark \\
  \midrule
  \multicolumn{3}{@{}l}{\textit{Hindsight Language Labeling}} \\ 
  BLOCKS \cite{bisk2016natural,bisk2018learning} & {\color{SimColor}30} & \textcolor{gray}{n/r} & \xmark  & \xmark & \cmark \\
  LangLFP \cite{lynch2020language}  & {\color{SimColor}10} & \textcolor{gray}{n/r} & \cmark   & \xmark & \xmark\\
  LOREL \cite{wu2021exampledriven,nair2022learning}   & {\color{RealColor}6}  & {\color{RealColor}1.7} & \cmark   & \cmark & \cmark\\
  CALVIN \cite{mees2022calvin}  & {\color{SimColor}20} & {\color{SimColor}0.4} & \cmark   & \xmark & \cmark \\
  \textbf{Language-Table} & \textbf{594} & \textbf{198} & \cmark & \cmark & \cmark \\
  \ \ \ \  ({\color{RealColor}\textit{real}}+{\color{SimColor}\textit{sim}})             & ({\color{RealColor}\textit{413}}+{\color{SimColor}\textit{181}}) & ({\color{RealColor}\textit{119}}+{\color{SimColor}\textit{79}})  \\
  \bottomrule
  \end{tabular}
  \caption{\textbf{Comparison of human-guided, language-labeled trajectory datasets}. Highlighted are the number of language-labeled trajectories and number of unique language labels (k=thousands) in {\color{RealColor}real} and {\color{SimColor}sim}, along with whether the data uses physical actions, real-world data, and if it is publicly available. \textcolor{gray}{n/r} means not reported.}
  \label{tab:data-comparison}
  \vspace{-3mm}
\end{table}

\textbf{Environment and Benchmark.}
Language-Table's simulated environment resembles our real-world tabletop manipulation scenario, which
consists of an xArm6 robot, constrained to move in a 2D plane with a cylindrical end-effector as in \cite{florence2022implicit}, in front of a smooth wooden board with a fixed set of 8 plastic blocks, comprising 4 colors and 6 shapes (Fig.~\ref{fig:new_capabilities}). 
In both simulation and real collection, we use high-rate human teleoperation with a 3rd person view (line-of-sight in real).
Actions are 2D delta Cartesian setpoints, from the previous setpoint to the new one. We batch collected training and inference data to 5hz observations and actions.
The Language-Table benchmark computes automated metrics for 5 task families, with 696 unique task variations. In addition to thresholded task success, a metric we find that better correlates with human-preferred performance is Success weighted by Path Length (SPL) \cite{anderson2018evaluation}, which trades off success rate against the efficiency of the path it took to succeed.
We note that policy hyperparameters ordered by SPL in \envname have thus far been ordered similarly in real-world performance. 
This provides a degree of validation for the simulated benchmark's relevancy to real world robotics.

\section{Policy Results and Discussion}
We present experiments aimed at answering the following questions:
(1) How capably can the system follow a wide variety of short-horizon natural language conditioned commands? 
(2) How capably can these skills be composed through interactive language guiding to accomplish a wide variety of multi-step long-horizon compositional rearrangements? 
(3) What is the benefit of being able to provide \textit{interactive} language feedback, compared to open-loop language plans? 
(4) Can one operator simultaneously guide several robots equipped with our policy? 
(5) Ablations: How does our transformer-based policy architecture compare to an existing visuo-linguo-motor baseline? How does our presented approach scale with varying amounts of data?

\subsection{Real world: diverse short-horizon language conditionable skills}
\begin{figure}[]
    \noindent
    \centering
    \includegraphics[width=\linewidth]{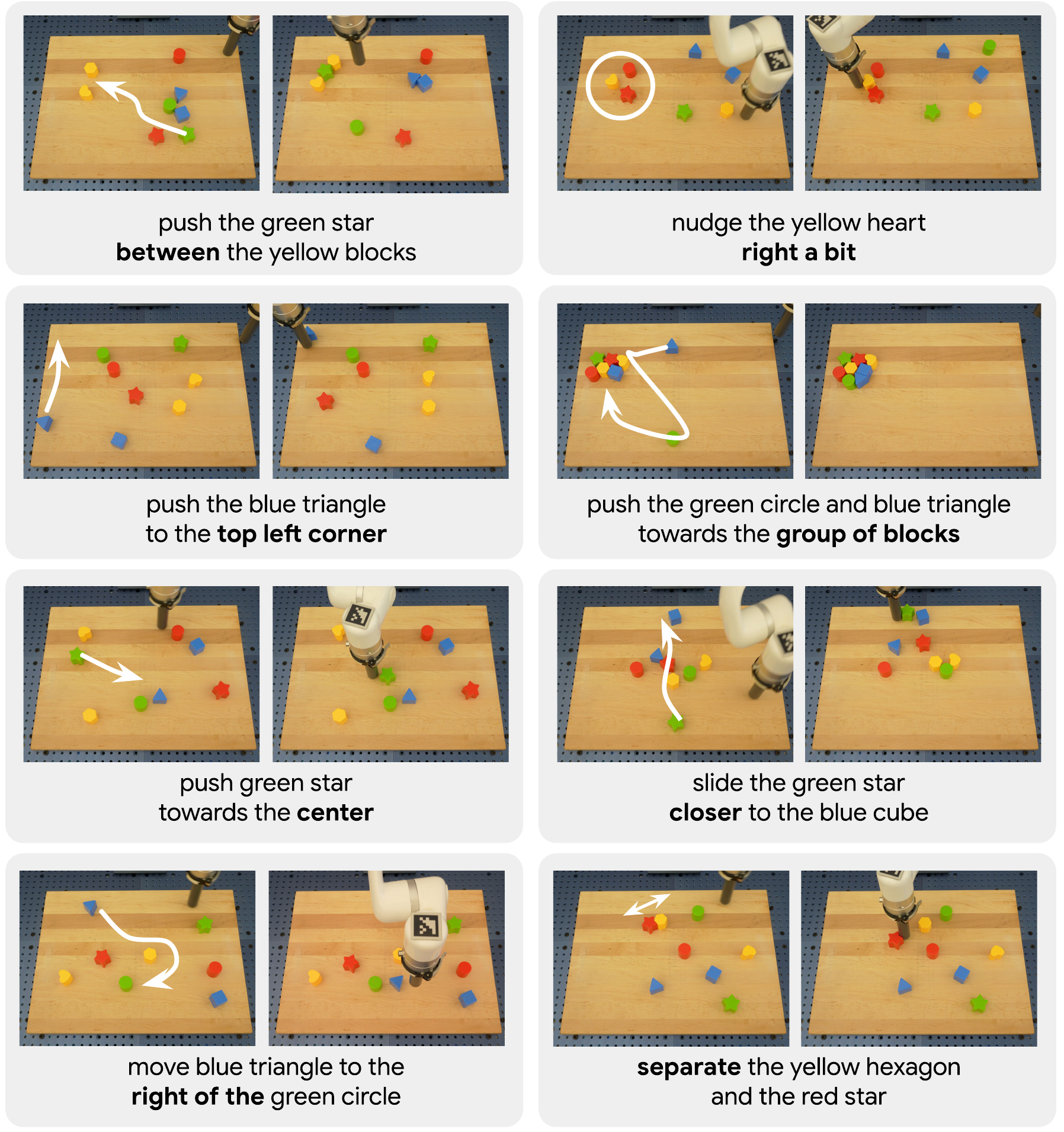}
    \caption{\textbf{Learning a wide variety of short-horizon open vocabulary behaviors.} \algname rollouts on a sample of the $>$87,000 crowdsourced natural language instructions we evaluate.}
    \label{fig:short_horizon_behaviors}
    \vspace{-8mm}
\end{figure}

Ideally, we would be able to evaluate an \algname policy on \textit{any} short-horizon command a real human might give it, which is intractable in general. As a surrogate, we estimate a $95\%$ confidence interval on average success over the 87,588 unique language instructions collected via crowdsourcing (20 randomly selected instructions, 10 trials each) available at time of analysis (Table~\ref{tab:data-stats}).
To succeed, policies must ground object properties and compositional spatial concepts (e.g. ``...\textit{top right side} of the \textit{yellow hexagon}" vs ``\textit{top right side} of the board"), and resolve difficult ambiguities (e.g. ``\textit{nudge} the cube left \textit{a bit}"). 
We report results in Table~\ref{tab:short-horizon}, with examples in Figure~\ref{fig:short_horizon_behaviors}. We see that \algname obtains a $93.5\%$ expected average success rate over all 87,588 instructions, $95\%$ CI $[90.08\%, 96.92\%]$. 
To our knowledge, this is the largest set of language conditioned behaviors a real-world policy has been shown to capably address, demonstrating a solid base capacity for language conditioned visuomotor control.

\begin{table}[]
\begin{tabular}{|l|l|}
\hline
\textbf{Short-Horizon Instruction}                          & \textbf{Success}           \\ \hline
(87k more...)                                  & ...                        \\ \hline
push the blue triangle to the top left corner & 80.0\%                     \\ \hline
separate the red star and the red circle      & 100.0\%                    \\ \hline
nudge the yellow heart a bit right            & 80.0\%                     \\ \hline
place the red star above the blue cube        & 90.0\%                     \\ \hline
point your arm at the blue triangle           & 100\%                      \\ \hline
push the group of blocks left a bit           & 100\%                      \\ \hline
\textbf{Average over 87k, CI 95\%}            & \textbf{93.50\% +- 3.42\%} \\ \hline
\end{tabular}
\caption{\textbf{Real world: Evaluating a wide variety of short-horizon language conditionable skills}. 95\% Confidence interval on the average success of our single policy over 87k (Table~\ref{tab:data-stats}) unique natural language instructions.}
\label{tab:short-horizon}
\vspace{-2mm}
\end{table}

\begin{figure*}[h!]
\centering
\hbox{\hspace{-0.7em} 
\includegraphics[scale=1.,width=1.01\textwidth]{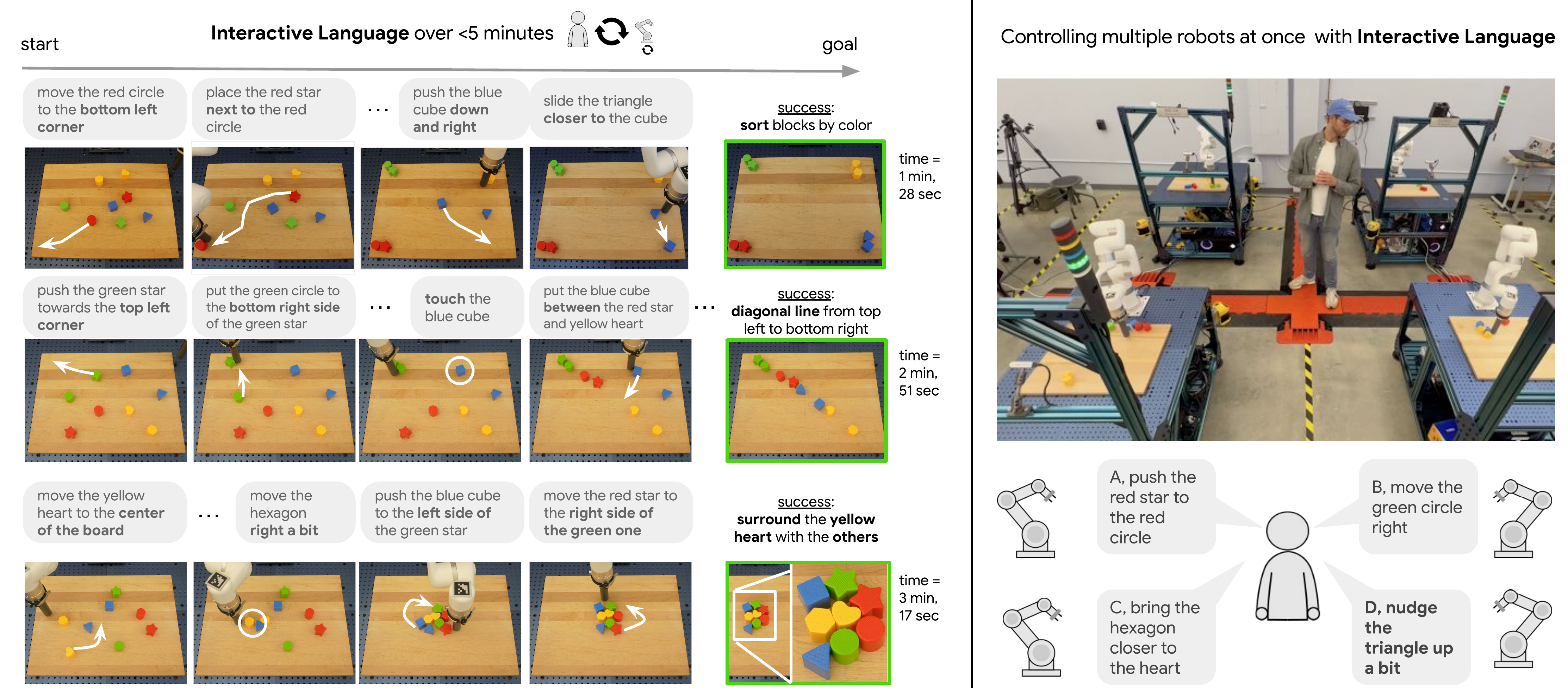}}
    \caption{\textbf{Capabilities explored with \algname}.
    Left: \textbf{Long-horizon language guidance} allows a human to guide a single policy to achieve a wide variety of long-horizon precise rearrangement goals. Language is used to interject new subgoals on-the-fly, to offer real-time corrections of unsafe or undesirable behavior (e.g. ``move the block away from the edge"), or to constrain the motions of the agent (e.g. ``slide the triangle \textit{slowly} left"). We evaluate policies on 11 goal families spanning hundreds of thousands of tasks. Right: \textbf{ Simultaneous multi-robot control}.
    Real time language allows a single human operator to guide multiple robots at once through the same long-horizon task, without requiring undivided attention to any one robot.}
    \vspace{-3mm}
    \label{fig:new_capabilities}
\end{figure*}

\subsection{Real world: long-horizon real-time language guidance}
\input{long_horizon_v1}

\begin{table}[]
\begin{tabular}{|l|l|l|}
\hline
\textbf{\begin{tabular}[c]{@{}l@{}}Language interaction \\ style\end{tabular}} & \textbf{\begin{tabular}[c]{@{}l@{}}Average number of \\ instructions provided\end{tabular}} & \textbf{\begin{tabular}[c]{@{}l@{}}Long-horizon\\ success \%\end{tabular}} \\ \hline
Open-loop                                                                         & 6.5                                                                                         & 25.0\% +- 18.98\%                                                          \\ \hline
Real-time (ours)                                                                  & 15                                                                                          & \textbf{85.0\% +- 15.65\%}                                                 \\ \hline
\end{tabular}
\caption{\textbf{Real world: long-horizon goal reaching via real-time human language guidance}. 95\% Confidence interval on the average success of our single real-time policy over 11 families and 100k possible goals, as compared to an open-loop baseline.}
\label{tab:long-horizon}
\vspace{-4mm}
\end{table}

\textbf{Multi-robot control via spoken language}. Finally, we investigate a new competency afforded by \algname: simultaneous multi-robot control.
In Figure~\ref{fig:new_capabilities}, see video as well, we see that four robots equipped with \algname policies can be guided at the same time by one operator. 
This language guided multi-robot control is, as far as we know, a capability not yet demonstrated in the literature. 
Importantly, due to short-horizon skill competency, this shows that language can relax the assumption of undivided operator attention, which is common for prior ways of correcting online robot behavior \cite{rakita2018autonomous, spencer2020learning, bajcsy2018learning}.

\begin{figure}[]
    \noindent
    \centering
    \includegraphics[width=\linewidth]{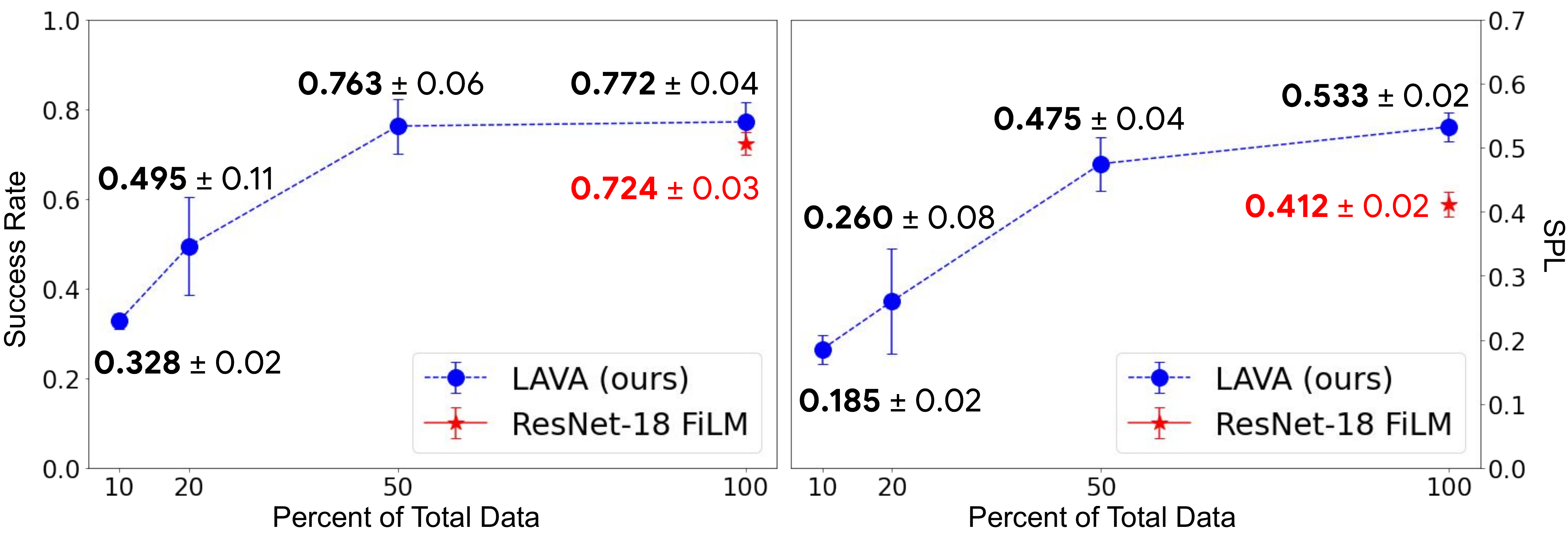}
    \caption{\textbf{Ablations in simulation.} We compare our LAVA transformer architecture to a baseline ResNet-18 FiLM model from \cite{jang2022bc}, as well as ablate the amount of data provided to training. We find the average success-weighted path length (SPL) to be a better indicator of qualitative performance than (unweighted) average success.}
    \label{fig:ablations}
    \vspace{-6mm}
\end{figure}


\subsection{Simulation: Architecture and data ablation}
In Figure~\ref{fig:ablations}, we present results in simulation ablating (i) our transformer-based policy architecture LAVA against the FiLM-conditioned ResNet architecture in \cite{jang2022bc} and (ii) the amount of data provided to policy training. We report average success and SPL \cite{anderson2018evaluation} over the multi-task benchmark in \envname (see ``Environment and Benchmark" in Section~\ref{sec:lang-table}), and all numbers are reported with confidence intervals over three seeded training runs. We see the presented architecture is indeed responsible for significant gains over prior work in SPL, a path-length-aware success metric we find correlates best with real world quality in our setup. When sweeping the amount of training data, we find that policy performance is seeing diminishing returns, but not yet plateauing across each doubling of data. While perhaps surprising given the scale of our collect, we believe that this result highlights the environment's complexity as well as the difficulty of open vocabulary visuomotor learning.

\section{Conclusion, Limitations, and Future Work}
We have presented and analyzed the \algname framework and we provide a number of associated assets, notably the \envname dataset and environment. 
We believe the scale of the dataset assets, the recipe used to produce them, the scale of the demonstrated policy diversity, and the exploration of new capabilities, each offer benefit to the research community in further advancing capable, realtime-conditionable visuo-linguo-motor robots.
While simple and scalable, our approach does have a number of limitations. 
The open problems in broader human-robot collaboration are numerous \cite{hayes2013challenges}, including intention detection, non-verbal communication, physically collaborative task completion, etc. Our approach addresses only the setting of real-time language-guided manipulation.
Future work may investigate applying \algname to important domains like real-time assistive robots, 
which may benefit from more capable natural language interfaces \cite{broad2016towards}.
We hope that our work can be useful as a basis for future research in capable, helpful robots with 
visuo-linguo-motor control.



\balance
\bibliographystyle{IEEEtran}
\bibliography{root}

\section*{Acknowledgements}

We would like to thank everyone who supported this research.  This includes robot teleoperators: Alex Luong, Armando Reyes, Elio Prado, Eric Tran, Gavin Gonzalez, Jodexty Therlonge, Joel Magpantay, Rochelle Dela Cruz, Samuel Wan, Sarah Nguyen, Scott Lehrer, Norine Rosales, Tran Pham, Kyle Gajadhar, Reece Mungal, and Nikauleene Andrews; robot hardware support and teleoperation coordination: Sean Snyder, Spencer Goodrich, Cameron Burns, Jorge Aldaco, Jonathan Vela; data operations and infrastructure: Muqthar Mohammad, Mitta Kumar, Arnab Bose, Wayne Gramlich; and the many who helped provide language labeling of the datasets. We would also like to thank Pierre Sermanet, Debidatta Dwibedi, Michael Ryoo, Brian Ichter and Vincent Vanhoucke for their invaluable advice and support.








\newpage
\input{appendix.tex}

\end{document}

%% file: v1_intro.tex
The goal of building a robot that can follow a diverse array of natural language instructions has been a longstanding goal of AI research, since at least the SHRDLU \cite{winograd1972understanding} experiments starting in the late 1960s.
While recent research on this topic has been abundant \cite{branavan2009reinforcement, tellex2011understanding, chen2011learning, matuszek2013learning,hill2019emergent,shridhar2022cliport,nair2022learning,jang2022bc},
few efforts have actually produced a robot that (i) exists in the real world, and (ii) 
can capably respond to a large number of rich, diverse language commands.
We expect that future research will continue to produce larger and more diverse sets of behaviors, either by sequencing raw skills together \cite{ahn2022can} or growing the number of raw skills themselves \cite{lynch2020language}.
However, we are also interested in (iii), the capacity to follow \textit{interactive} language commands, by which we mean that the robot reacts capably and in-the-moment to new natural language instructions provided during ongoing task execution. 
\begin{figure}[t]
    \vspace{-1.7mm}
    \noindent
    \centering
    \includegraphics[width=\linewidth]{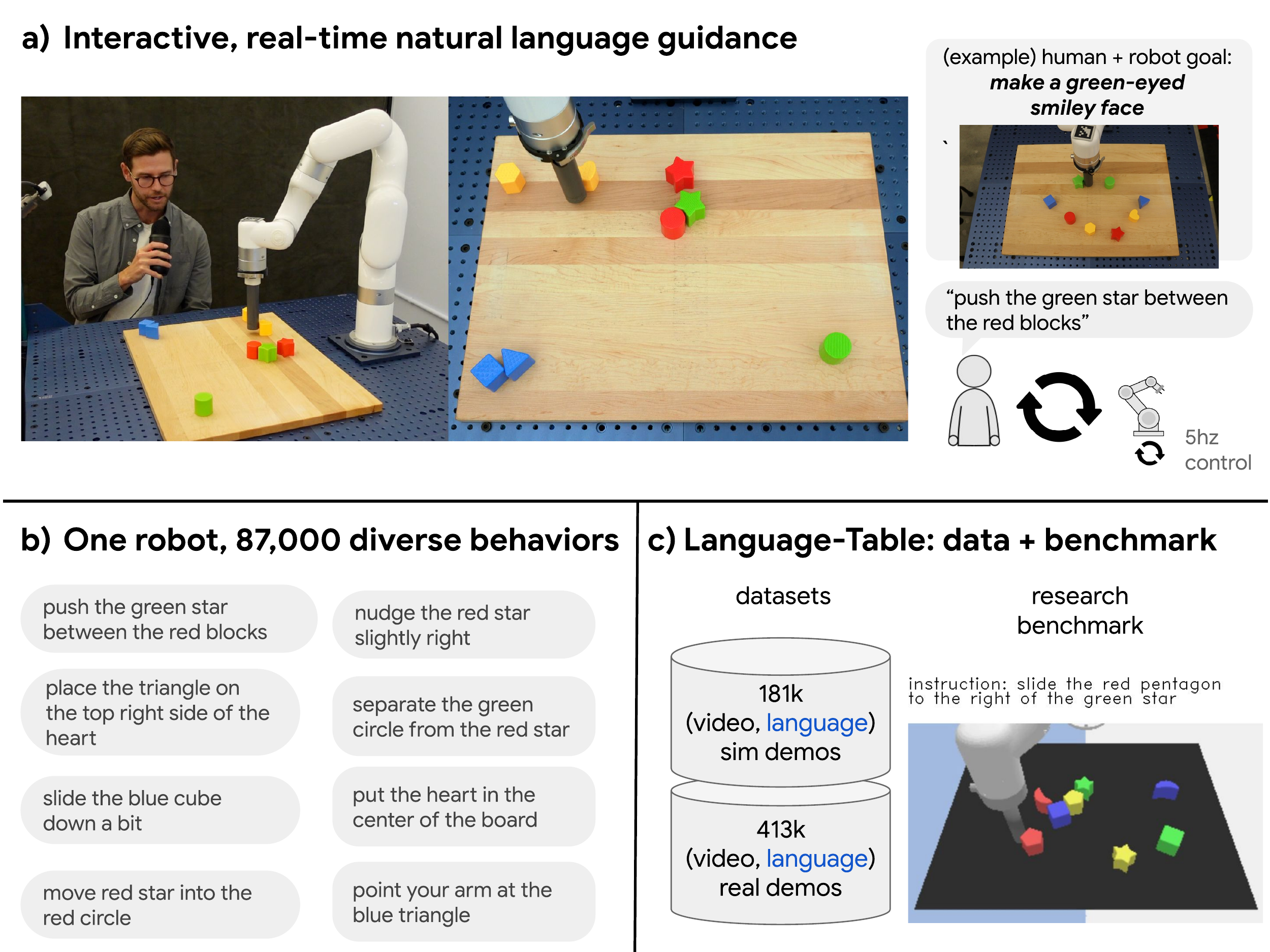}
    \caption{\textbf{Real-time language, diverse robot behaviors.}
    a) Over the course of 5 minutes, a human guides a robot to precisely rearrange objects a table into a desired shape, with real-time natural language as the only mechanism for specifying behaviors. b) We demonstrate a single robot that can capably address 87,000 behaviors specified entirely in natural language. c) We release Language-Table, a suite of human-collected datasets and a multi-task continuous control benchmark for open vocabulary visuolinguomotor learning.}
    \label{fig:moneyshot}
    \vspace{-9mm}
\end{figure}
Although we might expect such a robot to be possible given current methods, natural language-interactable robots are frequently slow in practice, and often use blocking parameterized skills \cite{shridhar2022cliport, ahn2022can} or simplifying self-resetting behaviors \cite{stepputtis2020language, jang2022bc} that prohibit this kind of live, real-time interaction.

In this paper, we demonstrate a framework for producing real-world, real-time-interactable, natural-language-instructable robots 
(Fig.~\ref{fig:moneyshot}, a) that by certain metrics operate at an order of magnitude larger scale than prior works.
To accelerate further research in this setting, we accordingly provide our associated recipe, dataset, models, hardware environment description, simulated analogue environment, and a research benchmark for language conditioned manipulation (Fig.~\ref{fig:moneyshot}, c).
In terms of scale, the produced robot policies can address 87,000 unique commands at an estimated 93.5\% success rate (Fig.~\ref{fig:moneyshot}, b), with 
continuous 5Hz
visuolinguomotor control, and are capable of chaining raw skills to reach hundreds of thousands of long horizon goals in its environment.
This robot exists in an environment which we designed to provide a tractable yet difficult level of challenge (perception from pixels, feedback-rich control, multiple objects, ambiguous natural language instructions).
We cast real time language guidance as a large scale imitation learning problem \cite{Pomerleau_behavior_cloning,osa2018algorithmic,lynch2020language} (Figure~\ref{fig:system}).
The learning algorithm recipe itself is intentionally simple, and instead the complexity of this effort was primarily in the data effort itself, for which we detail insights and techniques. We hope the dataset and benchmark may catalyze further work which may improve on our demonstrated sample complexity and performance.

Beyond demonstrating diverse short-horizon skills, we also use these capabilities to study the nonobvious benefits of a real-time language robot. For one, we show that through occasional human natural-language feedback, the robot can accomplish complex long-horizon rearrangements such as {\em{``put the blocks into a smiley face with green eyes''}} that require multiple minutes of precise coordinated control (Figure~\ref{fig:new_capabilities}, left). We also find that real-time language competency unlocks new capabilities like simultaneous, multi-robot instruction -- in which a single human can guide multiple real-time robots through long-horizon tasks (Figure~\ref{fig:new_capabilities}, right).

%% file: long_horizon_v1.tex
\textbf{Long horizon goal reaching}.
Next we aim to see whether humans can guide \algname policies through a wide range of multi-step compositional rearrangements.
We define over 100,000 language-distinct compositional goal states on our tabletop from 11 high level families (e.g. make high-level shapes, sort by color, place all blocks in specific locations, arrange into lines, etc.), then sample 20 uniformly from all 11. See Figure~\ref{fig:new_capabilities} for examples of different goal states. We evaluate each long horizon goal 3 times from randomly reset board states, yielding 60 total evaluations of a single policy.
We report success rates in Table~\ref{tab:long-horizon}.
We see that our policy obtains an $85.0\%$ expected average success rate on this diverse set of goals, $95\% \; CI \; [69.35\%, 100.00\%]$.
These results are best appreciated by watching the supplementary videos.
We note that reaching precise long horizon goals in the real world for even a single goal is a notoriously difficult problem for learning robots \cite{gupta2019relay}.
Even though our policies do not do so fully autonomously, we believe the fact that a real robot can address such a large and varied set of goals with real-time language feedback suggests a synergistic mode of future operation (at least until large improvements are made in the fully autonomous setting): robots learn a set of general-purpose low-level skills, and humans put them together in a familiar way using natural language, interrupting at any time to offer situation-specific corrections.

\textbf{Open-loop vs real-time language feedback}.
Next, we attempt to quantify the benefit of being able to provide \textit{real-time} language feedback, over the more common ``open-loop" evaluation setting where the sequence of subgoals is decided up front \cite{gupta2019relay, stepputtis2020language,lynch2020language, ahn2022can}.
We hypothesize that many of the tasks in our environment might require several rounds of iterative and interactive specification, due to the stochastic nature of single-point-of-contact pushing.
We perform the same evaluation as in the previous section, but the human operator commits up front to the set and order of commands they will provide.
We present results for this ablation in Table~\ref{tab:long-horizon}, finding that performance deteriorates from 85\% to 25\% when real-time language is removed. This indicates that for contact-rich tasks like the ones studied in this work, success depends heavily on sufficient \textit{real-time feedback}---not only for the low-level policy, but also for the agent providing it instructions.

%% file: appendix.tex


\newpage
\appendix

\subsection{Additional real-world experiment details}
\label{sec:appendix_real_detail}
Our real-world experiments use UFACTORY xArm6 robot arms with all state logged at 100 Hz. Observations are recorded from an Intel RealSense D415 camera, using RGB-only images at 640x360 resolution, logged at 30 Hz, which we resize to 320x180 before handing to robot policies.
Policies use 320x180 single-camera RGB-only images, with no other observations besides language. The asynchronous observations and actions are batched to psuedo-synchronous 5 Hz pairs for training the policy, with camera latency (characterized at roughly 80 ms) accounted for when forming psuedo-synchronous training pairs.
The cylindrical end-effector is made from a 6 inch long plastic PVC pipe sourced from McMaster-Carr (\href{https://www.mcmaster.com/9173K515/}{9173K515}). The work surface is 24 x 18 inch smooth wood cutting board. The manipulated objects are from the Play22 Baby Blocks Shape Sorter toy kit (\href{https://play22usa.com/shop/ols/products/16olfxvr5t}{Play22}). The 6DOF robot is constrained to move in a 2D plane above the table.

\subsection{Language-Table: Datasets}
Here we outline the various datasets available in Language-Table, across simulation and real.

\subsubsection{Simulation-Raw-Collect}
This dataset consists of 6 teleoperators teleoperating a robot in simulation, following long horizon prompts. See representative prompts in Table~\ref{tab:collection-prompts}. 8318 episodes were collected with an average length of 36.8 $\pm$15 seconds, yielding a total of 85.5 hours of raw data.

\subsubsection{Simulation-Relabeled} The Simulation-Raw-Collect data was sent to 64 crowdsourced annotators, who used the interface described in Appendix~\ref{sec:appendix_event_selectable} to generate 181,020 hindsight relabeled trajectories, with 78,623 unique instructions. See representative instructions in Table~\ref{tab:representative-instructions}.

\subsubsection{Real-World-Raw-Collect} This dataset consists of 11 teleoperators alternating over four robots, following long horizon prompts. See representative prompts in Table~\ref{tab:collection-prompts}. 23498 total episodes were collected with an average length of 9.9 minutes $\pm$5.6 seconds, yielding a total of 3865 hours of raw data. Note that 16417 episodes totaling 2701 hours went into the actual training of policies, and the remaining was collected after training the demonstrated policy, but before releasing the dataset.

\subsubsection{Real-World-Relabeled} The Real-World-Raw-Collect data was sent to 64 crowdsourced annotators, who used the interface described in Appendix~\ref{sec:appendix_event_selectable} to generate 414,798 total hindsight relabeled trajectories, with 119,959 unique instructions. See representative instructions in Table~\ref{tab:representative-instructions}. Note that 298,782 relabeled trajectories went into training, with 87,140 unique instructions, and the remaining was collected post-training, but pre-release.

\subsection{Language-Table: Environment}
Our simulated environment is intended to roughly match our real world setup, and consists of a simulated 6DoF robot
xArm6 implemented in PyBullet \cite{coumans2016pybullet} equipped with a small cylindrical end effector. Third person perspective 320x180 RGB-only images from a simulated camera are used as visual input. On a board in front of the robot are 8 blocks: red crescent, red pentagon, blue crescent, blue cube, green cube, green star, yellow star, and yellow pentagon. Like in the real world, the arm is constrained to the 2D plane and the action space is the delta 2D cartesian setpoint of the end effector. We run all experiments from RGB and language input only, but the environment additionally exposes 26-dimensional state observations (2D position and 1D rotation angle for each block, 2D position of end effector). While our real world policies perform asynchronous inference and control at 5hz, the policies in Language-Table perform blocking control at 10hz. Despite this difference, and others like differences between real and simulated images, we found policy performance in Language-Table was highly correlated with policy performance in the real world.

\subsection{Language-Table: Evaluation}
We define five simulated evaluation task families (spanning 696 unique task conditions) in Language-Table, each with a hand-defined success criterion:
\begin{itemize}
    \item \textit{block2block}: Push a block to another block. Success is thresholded distance between source and target block. There are 56 unique task conditions (8 source blocks x 7 target blocks).
    \item \textit{block2abs}: Push a block to an absolute location on the board: top left, top center, top right, center left, center, center right, bottom left, bottom center, bottom right. Success is thresholded distance between block and target location. There are 72 unique task conditions (8 blocks x 9 locations).
    \item \textit{block2rel}: Push a block to a relative offset location: left, right, up, down, up and left, up and right, down and left, down and right. Success is the thresholded distance between the block and the invisible target offset location. There are 64 unique task conditions (8 blocks x 8 offset directions).
    \item \textit{block2blockrel}: Push a block to a relative offset location of another block: left side, right side, top side, bottom side, top left side, top right side, bottom left side, bottom right side. Success is the thresholded distance between the source block and the invisible target offset location of the target block. There are 448 unique task conditions (8 source blocks x 7 target blocks x 8 offset directions).
    \item \textit{separate}: Separate two blocks. Success is the thresholded distance between the two blocks. There are 56 unique task conditions (8 source blocks x 7 target blocks).
\end{itemize}
These task families were used to benchmark models in simulation, allowing us to find hyperparameters that transferred well to fully-real-world training (we note that there was no sim-to-real component in the training employed by this work). The language conditioning for the automated evaluation tasks are generated synthetically from predefined synonym sets for each task condition.

\subsection{Event selectable hindsight relabeling details}
\label{sec:appendix_event_selectable}
\begin{figure}[t]
    \noindent
    \centering
    \includegraphics[width=\linewidth]{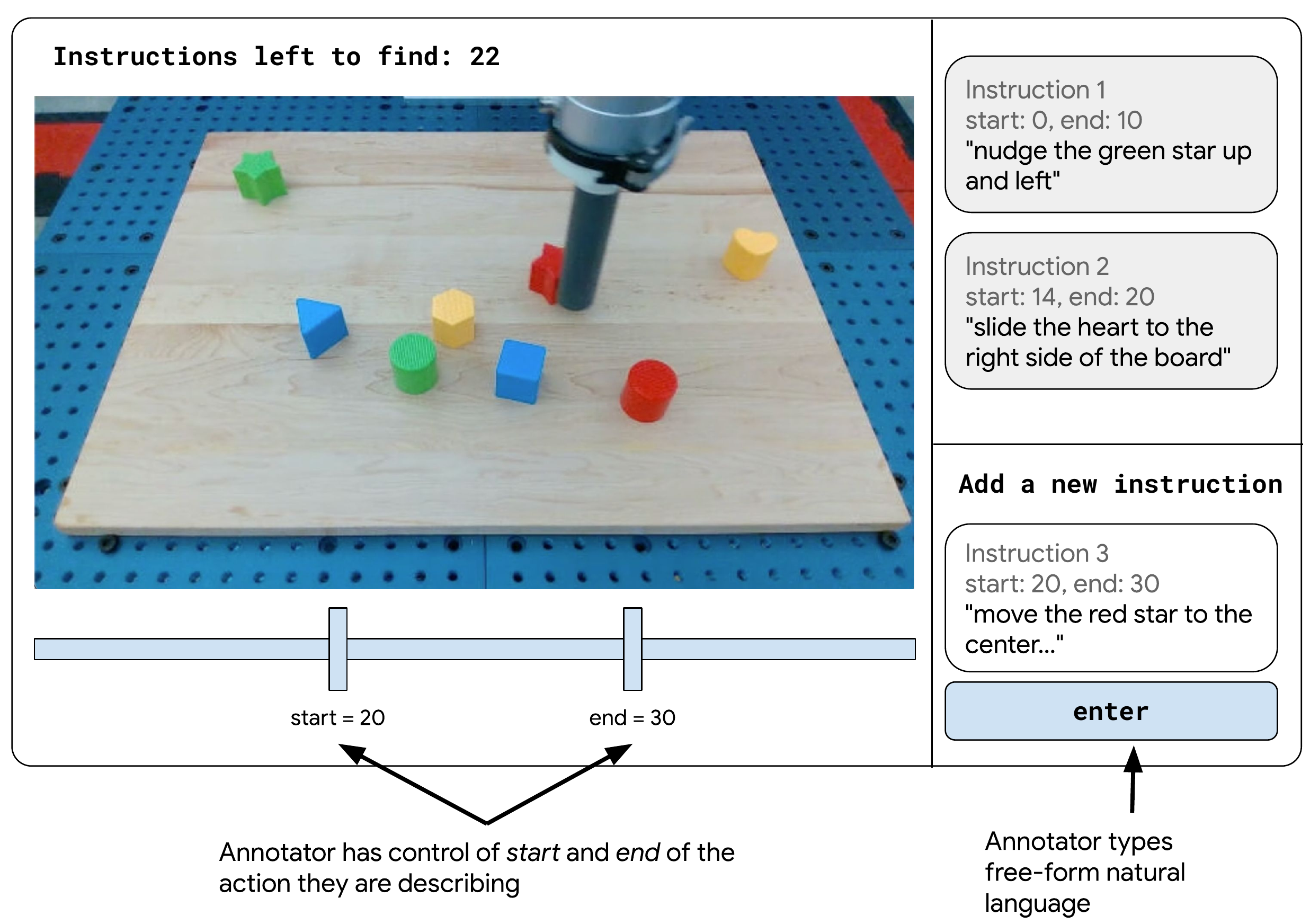}
    \caption{\textbf{Mockup of our event selectable hindsight relabeling interface.}}
    \label{fig:event_selectable}
\end{figure}

Figure~\ref{fig:event_selectable} depicts a mockup of the interface our crowdsourced workers used to do event selectable hindsight relabeling. We asked data labelers to first watch an entire long horizon video, then produce 12 medium horizon and 12 short horizon actions, where the definition is left to rater discretion. Labelers have control of temporal segmentation tools, allowing them to mark the beginning and end of each action, and they describe each action as an open vocabulary natural language instruction.

\subsection{Model architecture details}
Here we describe LAVA (``Language Attends to Vision for Actions"), the transformer-based visuo-linguo-motor neural network architecture we use in this work. Internally, our architecture consists of a perception module, language module, vision-language fusion module, temporal fusion module, and policy output. We describe each below.

\textbf{Perception module}. Each training example consists of $(s, a, l)_i \sim \DTRAINING$, where
$s \in \mathbb{R}^{\operatorname{seqlen}\times320\times180\times3}$ is RGB observation history, and for the shown policies we used $\operatorname{seqlen}$=4. We pass each frame in the video $s$ through a convnet to obtain multi-scale visual feature descriptors (features at multiple layers). Our convnet consists of two Imagenet-pretrained ResNet \cite{deng2009imagenet, he2016deep} layers and two additional learned convolutional layers with channel sizes 128 and 256 respectively and 2D max pooling between each layer. This yields a multi-scale feature pyramid for each image with $[H, W]$ of sizes [[112, 112], [56, 56], [28, 28], [14, 14], [7, 7]].

\textbf{Language module}. We use a pretrained CLIP text encoder \cite{radford2021learning}, which is finetuned on our in-domain data, but remains fixed during policy training. We use a simple contrastive method for finetuning models pretrained  on  (image,  language) pairs to domains where the observations are (video, language) pairs: generate (start frame $s_0$, goal frame $s_g$, language $l$) from all videos, and then during finetuning, pass concatenated image encodings $\operatorname{concat}([z_0, z_g])$ through an MLP to get a single encoding $z^{\operatorname{im}}$ with the same dimensionality as encoded language $z^{\operatorname{lang}}$. We preprocess text by stripping punctuation and extra spaces, but apply no additional preprocessing or augmentation. Cleaned text is passed through the CLIP embedder to get a sentence embedding with dimensionality 512.

\textbf{Vision-Language Fusion Module}. We fuse visual and lingual information using a ``Language-Attends-to-Vision" transformer block.
For a single image position, this block takes as input (i) multi-scale pixel features (in our case the CNN features at zero-indexed layers 2, 3, 4 with $H,W$ sizes [28, 28], [14, 14], [7, 7]) and (ii) a sentence embedding (in our case the 512-dimensional CLIP-encoded $l$). First, we map each layer $n$ to $[H_n, W_n, \operatorname{dmodel}]$ using a layer-specific MLP, then 2D position encode each feature map with 2D sinusoidal positional embeddings. We then flatten all the multi-scale features into one long visual token list (in our case with shape [1029, $\operatorname{dmodel}$]). We project language to $\operatorname{dmodel}$ using an MLP, and apply dropout to both projected image and language features.
We then iteratively fuse vision and language features, handing the sentence token as query and visual tokens as keys and values to a standard pre-norm decoder-only transformer \cite{xiong2020layer} performing cross-attention, with only language on the residual path. Our vision-language transformer had 4 layers, with $\operatorname{dmodel}$=128, 2 heads, feed forward width of 128, and dropout of 0.1.

\textbf{Temporal Fusion Module.} The output from applying our vision-language fusion module to each image in the $seqlen=$4 context history is a $[\operatorname{seqlen}, \operatorname{dmodel}]$ sequence of vision-language embeddings. We apply 1D sinusoidal positional encoding to each element of the sequence, then feed the sequence to a standard pre-norm transformer performing self-attention, also outputting $[\operatorname{seqlen}, \operatorname{dmodel}]$, which we average pool over the time dimension. Our temporal transformer had 2 layers, with $\operatorname{dmodel}$=128, 2 heads, feed forward width of 128, and dropout of 0.1.

\textbf{Policy Output.} We hand the average-pooled $\operatorname{dmodel}$ embedding to a deep residual MLP with 2 blocks of residual width 1024. Each block has 3 MLP layers, the first two with width 256 and the final with width 1024. All MLPs have ReLU activation with normal initialization on kernel and bias. Finally we use a linear projection to the 2D action space. 

\subsection{Training details}
We train our policies on a TPUv3 8x8 pod (64 TPUv3 chips) for approximately 500,000 steps or until training loss plateaus. At roughly 7.6 steps/second, policies finish training in 18 hours. All models are trained with Adam \cite{kingma2014adam} with default TensorFlow momentum parameters, learning rate 1e-3, and a batch size of 4096. Action labels are normalized using statistics collected from training to $\mathcal{N}(0, 1)$.

\subsection{Ablations}

\begin{table}[]
\begin{tabular}{ll}
\hline
Method     & Success Rate \\ \hline
Full LAVA + training recipe (ours)  & \textbf{0.772} $\pm$ 0.044  \\
No CLIP Finetuning    & 0.735 $\pm$ 0.023  \\
No Temporal Fusion Module    & 0.732 $\pm$ 0.036  \\
Half Batch Size (2048) & 0.720 $\pm$ 0.038  \\ \hline
\end{tabular}
\caption{Results of experiments to ablate model architecture details in the Language-Table simulator. All results are reported over 3 seeds after 350k steps.}
\label{tab:ablation-results}
\end{table}

We ablate the following training details. The results are reported in Table \ref{tab:ablation-results}.
\begin{itemize}
  \item \textbf{CLIP Finetuning.} We evaluate the importance of finetuning the CLIP language module on in-domain data. We use the pretrained weights for the ViT-B/32 model from \cite{radford2021learning} to encode language instructions without finetuning the text encoder on any Language-Table data. The text encoder still remains fixed during training. Although this results in only a few percent drop in the Language-Table sim, we observe a much stronger qualitative difference in model behavior on the real robot. This difference between sim and real could be explained by the Language-Table sim evaluation using a fixed set of templated instructions, while the real world evaluation uses more diverse language from human operators.
  \item \textbf{Temporal Fusion Module.} We evaluate the importance of using our transformer-based temporal fusion module. We encode each image in the $seqlen=4$ context history by stacking the images channel-wise and encoding the images using a randomly initialized ConvNet. The multi-scale visual feature descriptors from this ConvNet are still fed into the "Language-Attends-to-Vision" transformer block, with no additional self-attention temporal fusion transformer.
  \item \textbf{Batch Size.} We evaluate the effect of batch size on our model by reducing our batch size in half to 2048. We see that this results in a performance drop after 350k steps.
\end{itemize}

\subsection{Extended Related Work}
Recent work has leveraged large language models (LLMs) to generate sequences of subgoals for language conditioned policies. These can be ``open-loop" \cite{huang2022language,ahn2022can}, which lack an mechanism for replanning, or ``closed-loop" \cite{huang2022inner}, which generate up-to-date plans by prepending textual descriptions of the current scene to the prompting of the LLM planner. A limitation of both formulations is that when tasks that call for fine-grained spatial detail (like those examined in this work), it is difficult for LLMs to generate accurate subgoals from purely textual scene descriptions. 
Although visual language models (VLMs) like \cite{alayrac2022flamingo} suggest a promising direction, matching human levels of perception and cognition to effectively guide policies towards arbitrary goals remains a difficult open challenge.
Our work is complementary in that our focus is instead on obtaining a large diverse set of short-horizon behaviors, and ones that can be interactively conditioned in real time. Combining autonomous long-horizon planning together with our demonstrated recipe for short-horizon behaviors is a strong candidate for future work.

\begin{table}[]
\begin{tabular}{|l|}
\hline
slide the green circle into the top side of the yellow hexagon        \\ \hline
slide the green star along with the yellow hexagon towards the center \\ \hline
move your arm near the bottom center                                  \\ \hline
push the yellow heart closer to the yellow hexagon and blue triangle  \\ \hline
move the blue triangle into group of blocks                           \\ \hline
push the red star upwards                                             \\ \hline
place the yellow heart to the left side of the green star             \\ \hline
place the green staryellow hexagon at the center of the board         \\ \hline
nudge red star along with red circle a bit up                         \\ \hline
move the group of blocks to the centre of the board                   \\ \hline
move the arm left beside the red star                                 \\ \hline
slide the blue triangle along with yellow hexagon slightly up         \\ \hline
move the blue cube towards the center                                 \\ \hline
push the red star along with the red circle towards the top center    \\ \hline
push red star below the yellow heart                                  \\ \hline
separate yellow hexagonn from the blue cube                           \\ \hline
move the red circle right and down a bit                              \\ \hline
slide the blue cube towards left                                      \\ \hline
move the blue triangle along with the red circle slightly right       \\ \hline
push the blue triangle to the bottom right of the blue cube           \\ \hline
\end{tabular}
\caption{Representative examples of crowdsourced instructions obtained via hindsight relabeling.}
\label{tab:representative-instructions}
\end{table}

\begin{table}[]
\begin{tabular}{|l|}
\hline
put all the blocks in a vertical line on the right of the board                                                                                                                                                                                                                                                                \\ \hline
group the blocks by color                                                                                                                                                                                                                                                                                                      \\ \hline
\begin{tabular}[c]{@{}l@{}}make one horizontal line out of the red and blue blocks, \\ then make a horizontal line out of the green and yellow blocks\end{tabular}                                                                                                                                                             \\ \hline
\begin{tabular}[c]{@{}l@{}}make one horizontal line out of the blue and green blocks, \\ then make a horizontal line out of the red and yellow blocks\end{tabular}                                                                                                                                                             \\ \hline
\begin{tabular}[c]{@{}l@{}}put the: \\ 0) green circle to top left, \\ 1) red circle to top center, \\ 2) green star to top right, \\ 3) red star to center left, \\ 4) blue triangle to center right, \\ 5) yellow heart to bottom left, \\ 6) yellow hexagon to bottom center, \\ 7) blue cube to bottom right,\end{tabular} \\ \hline
put 3 blocks in the bottom left corner, then the rest in the center left                                                                                                                                                                                                                                                       \\ \hline
\begin{tabular}[c]{@{}l@{}}make one horizontal line out of the red and green blocks, \\ then make a vertical line out of the blue and yellow blocks\end{tabular}                                                                                                                                                               \\ \hline
put the blocks in a diagonal line from the top left to bottom right                                                                                                                                                                                                                                                            \\ \hline
\begin{tabular}[c]{@{}l@{}}put the yellow and red blocks together in a group,\\ then put the green and blue blocks together in a group\end{tabular}                                                                                                                                                                            \\ \hline
\begin{tabular}[c]{@{}l@{}}put the blue and green blocks together in the bottom center,\\ then put the red and yellow blocks together in the center right\end{tabular}                                                                                                                                                         \\ \hline
\begin{tabular}[c]{@{}l@{}}put the: \\ 0) blue triangle to top left, \\ 1) yellow hexagon to top center, \\ 2) green star to top right, \\ 3) blue cube to center left, \\ 4) red circle to center right, \\ 5) red star to bottom left, \\ 6) yellow heart to bottom center, \\ 7) green circle to bottom right\end{tabular}  \\ \hline
surround the yellow heart with the others                                                                                                                                                                                                                                                                                      \\ \hline
put all the blocks in the bottom right corner                                                                                                                                                                                                                                                                                  \\ \hline
make a "V"" shape out of all the blocks                                                                                                                                                                                                                                                                                        \\ \hline
\begin{tabular}[c]{@{}l@{}}put the red blocks in the center right, the yellow blocks in the bottom center, \\ the green blocks in the bottom right corner, and the blue blocks in the top center\end{tabular}                                                                                                                  \\ \hline
\begin{tabular}[c]{@{}l@{}}put the: \\ 0) blue cube to top left, \\ 1) yellow heart to top center, \\ 2) blue triangle to top right, \\ 3) red star to center left, \\ 4) yellow hexagon to center right, \\ 5) green star to bottom left, \\ 6) green circle to bottom center, \\ 7) red circle to bottom right\end{tabular}  \\ \hline
\begin{tabular}[c]{@{}l@{}}put the: \\ 0) green circle to top left, \\ 1) yellow heart to top center, \\ 2) blue cube to top right, \\ 3) red circle to center left, \\ 4) blue triangle to center right, \\ 5) green star to bottom left, \\ 6) red star to bottom center, \\ 7) yellow hexagon to bottom right\end{tabular}  \\ \hline
put all the blocks in the top center                                                                                                                                                                                                                                                                                           \\ \hline
\begin{tabular}[c]{@{}l@{}}put the red blocks in the top right corner, the green blocks in the center right, \\ the blue blocks in the bottom center, and the yellow blocks in the center\end{tabular}                                                                                                                         \\ \hline
put all the blocks in a vertical line on the center of the board                                                                                                                                                                                                                                                               \\ \hline
\end{tabular}
\caption{Representative examples of the prompts used to drive collection. These are discarded after collection.}
\label{tab:collection-prompts}
\end{table}
